\documentclass[conference]{IEEEtran}
\IEEEoverridecommandlockouts
\usepackage{amsmath,amssymb,amsfonts}
\usepackage{algorithmic}
\usepackage{graphicx}
\usepackage{textcomp}
\usepackage{xcolor}
\usepackage{graphicx}%
\usepackage{multirow}%
\usepackage{amsmath,amssymb,amsfonts}%
\usepackage{amsthm}%
\usepackage{mathrsfs}%
\usepackage{xcolor}%
\usepackage{textcomp}%
\usepackage{manyfoot}%
\usepackage{booktabs}%
\usepackage{listings}%
\usepackage{cuted}
\usepackage{tikz}

\def\BibTeX{{\rm B\kern-.05em{\sc i\kern-.025em b}\kern-.08em
    T\kern-.1667em\lower.7ex\hbox{E}\kern-.125emX}}
\begin{document}


\title{Modular Transformer Architecture for \\Precision Agriculture Imaging}

\author{\IEEEauthorblockN{1\textsuperscript{st} Brian Gopalan}
\IEEEauthorblockA{\textit{University Park} \\
\textit{The Pennsylvania State University}\\
PA, United States \\
briang@psu.edu}
\and
\IEEEauthorblockN{2\textsuperscript{nd} Nathalia Nascimento}
\IEEEauthorblockA{\textit{Engineering Division, Great Valley} \\
\textit{The Pennsylvania State University}\\
PA, United States \\
nnascimento@psu.edu}
\and
\IEEEauthorblockN{3\textsuperscript{rd} Vishal Monga}
\IEEEauthorblockA{\textit{University Park} \\
\textit{The Pennsylvania State University}\\
PA, United States \\
vum4@psu.edu}
}

\maketitle

\begin{abstract}
This paper addresses the critical need for efficient and accurate weed segmentation from drone video in precision agriculture. 
A quality-aware modular deep-learning framework is proposed that addresses common image degradation by analyzing quality conditions—such as blur and noise—and routing inputs through specialized pre-processing and transformer models optimized for each degradation type. The system first analyzes drone images for noise and blur using Mean Absolute Deviation and the Laplacian. Data is then dynamically routed to one of three vision transformer models: a baseline for clean images, a modified transformer with Fisher Vector encoding for noise reduction, or another with an unrolled Lucy-Richardson decoder to correct blur. This novel routing strategy allows the system to outperform existing CNN-based methods in both segmentation quality and computational efficiency, demonstrating a significant advancement in deep-learning applications for agriculture.
\end{abstract}


\section{Introduction}\label{sec1}
AI-powered computer vision is becoming essential for precision agriculture, with drone-based systems providing critical data for tasks like automated weed identification. A major challenge for these systems is image degradation from drone-induced noise and motion blur. Although traditional CNN-based solutions are often used for denoising and deblurring, more powerful transformer architectures have proven to be more effective. However, this superior performance comes at a cost, as transformers require significantly more computational resources. In addition, recent studies have shown that even transformer-based models exhibit varying degrees of sensitivity to different types of image degradation, particularly noise and blur, which can substantially impact their performance in real-world scenarios \cite{varga2024understanding}. To address this performance-efficiency trade-off, a quality-aware modular deep-learning framework is proposed. This approach provides the flexibility to select and apply the most appropriate models on a per-image basis, allowing for high-quality segmentation while maintaining computational efficiency.

This paper presents a modular deep learning approach with routing and shared computational functions to address noise and blur in drone footage for performing weed segmentation. The contributions of this paper include:

\begin{enumerate}
    \item A modular deep learning approach that encompasses data routing and computational functions to optimize efficiency and select models based on dynamic factors. The routing function utilizes a threshold for the Laplacian of the image to route information to the deblur module; and a threshold for the Mean Absolute Deviation of the image’s high-pass filtered coefficients to route information to the denoise module.
    \item Modification of a vanilla vision transformer for addressing blur by incorporating an unrolled Lucy-Richardson algorithm into the transformer decoder.
    \item Modification of a vanilla vision transformer for addressing noise by employing Fisher Vectors instead of patches in the transformer encoder.
\end{enumerate}

These contributions collectively enable weed segmentation from real-world degraded drone imagery using the Sorghum dataset \cite{genze_improved_2023}, which contains 1,300 blurred–sharp image pairs captured over a sorghum field.

\section{Background}

Modular deep learning \cite{pfeiffer_modular_2023} is a powerful approach that aims to solve two key challenges in traditional deep learning: computational efficiency and generalizability. Unlike a standard, monolithic pipeline where all data is processed by the same set of operations, this framework uses a router to intelligently route incoming data to the most relevant components. This selective processing ensures that computational resources are used efficiently by only activating the modules required for a specific task.
The modular design also has significant benefits for development and collaboration. By breaking down the model into reusable components, the architecture becomes inherently more flexible and easier to extend. This makes it simpler for researchers and developers to add new functionalities or adapt the system for different applications. An extension of this modular concept is dynamic model selection \cite{nascimento2018context}. This framework takes the idea of routing a step further by using context-aware factors to choose the most appropriate model from a collection of pre-trained options. Following the definition by Abowd et al. \cite{abowd1999towards}, we consider image quality—specifically degradation factors like blur and noise—as contextual information that characterizes the state of the input image. This context guides dynamic model selection in our framework.

The growing prevalence of drones in agriculture, particularly for tasks like automated weeding, has brought to light specific challenges inherent in aerial imagery. Two primary issues are common in these images and significantly impact subsequent analysis \cite{cavaliere_semantically_2019}. First, agricultural fields are often characterized by noise, which can be introduced by physical conditions such as dust and other particulates. Second, the movement of the drone itself during image capture can introduce motion blur. Both noise and blur degrade image quality, thereby hindering the performance of critical downstream tasks, such as the segmentation required to accurately identify the location of weeds.

\section{Related Work}


Machine learning (ML) has emerged as a key driver in the agricultural sector \cite{liakos_machine_2018}. Specifically, ML algorithms have made significant contributions in areas such as crop management for yield prediction and disease and weed detection. In an in-depth study, Rejeb et al. \cite{rejeb_drones_2022} conducted a bibliometric analysis and discovered a growing number of publications on the utilization of drones in agriculture. Notably, the drone market in the United States for agricultural purposes has reached a substantial value of USD 841.9 million. Recent examples of ML applications in agriculture include Van Essen et al. \cite{van_essen_uav-based_2025}, who employed a reinforcement learning-based approach to optimize the route taken by a drone for weed detection. Similarly, Khuimphukhieo et al. \cite{khuimphukhieo_estimating_2025} utilized drone footage to predict the yield from sugarcane fields using a random forest-based approach. Deep learning-based computer vision algorithms have made a substantial impact in weed detection by utilizing multispectral images captured from drones \cite{pantazi_active_2016}. However, as Cavaliere et al. \cite{cavaliere_semantically_2019} point out, videos taken from drones are often degraded by noise and blur, which negatively affects the performance of computer vision tasks.

The solution space for addressing noise and blur encompasses ML techniques such as convolutional neural networks (CNNs). Lecun et al. \cite{lecun_gradient-based_1998} introduced a pioneering CNN architecture for handwriting recognition, which became the foundation for decades of widespread use of CNNs in computer vision. CNNs have proven effective in deblurring images. Genze et al. \cite{genze_improved_2023} employed a combination of two CNN-based architectures—UNET and NAFNET—to remove blur from drone images, enabling weed segmentation. Similarly, Hou et al. \cite{hou2025qmix} proposed QMix, a quality-aware CNN-based model that modulates internal feature processing based on input noise levels. While their work focuses on retinal disease classification, it highlights the potential of incorporating quality-awareness into deep learning pipelines for robust prediction under image degradation.

As deep learning gained popularity, traditional computer vision algorithms have been adapted to leverage this paradigm. This includes the authors’ \cite{chen_spatio-temporal_2013} innovation to introduce Fisher Vector encoding as part of their video classifier architecture, which effectively reduced noise. Within the context of the transformer architecture, the sole known application of Fisher Vectors pertains to natural language processing (NLP) for language identification \cite{krebbers2022multi}. Similarly, authors \cite{chen_deep_2023} introduced the use of the Lucy-Richardson algorithm for deconvolution, utilizing the CNN architecture.

While CNN-based solutions have demonstrated success in computer vision, the transformer architecture, initially introduced by Vaswani et al. \cite{vaswani_attention_2023} for natural language processing (NLP) problems, has gained popularity. Dosovitskiy et al. \cite{dosovitskiy_image_2021} repurposed the transformer architecture for computer vision tasks. Another recently proposed framework is modular deep learning \cite{pfeiffer_modular_2023}. This framework modularizes computational, routing, and aggregation functionalities to enhance efficiency by eliminating redundancies and facilitating the extensibility of deep learning solutions. Additionally, the modular framework enables task generalization and efficient transfer learning. 

Extending the modular paradigm, Nascimento et al. \cite{nascimento2018context} introduced a framework for selecting models based on dynamic factors. Their solution used simple neural networks and was applied to non-visual, structured sensor data. In contrast, we design a novel modular and context-aware framework specifically tailored for visual processing with transformer-based models, which are known to exhibit varying performance under different image quality conditions \cite{varga2024understanding}. Our approach is not a direct application of previous architectures but a new design that integrates routing and quality-aware specialization to address degradation issues in drone imagery—a complex and high-variance visual domain.





\section{Proposed Approach}
The proposed approach efficiently routes images to one of three models based on their characteristics. These three models are implemented in a modular manner to facilitate this process. The three modules involved are:
\begin{enumerate}
  \item Vision transformer module (ViT)
  \item Vision transformer with modular routing to address noise using Fisher Vector encoding (FV)
  \item Vision transformer with modular routing to address blur using Lucy-Richardson algorithm (LR)
\end{enumerate}
The architecture also enables combinations of these modules. For instance, if both noise and blur are present, the router will select both Fisher Vector encoding and Lucy-Richardson deblurring techniques. 

\begin{figure*}[htbp]
    \centering
    \includegraphics[width=0.8\linewidth]{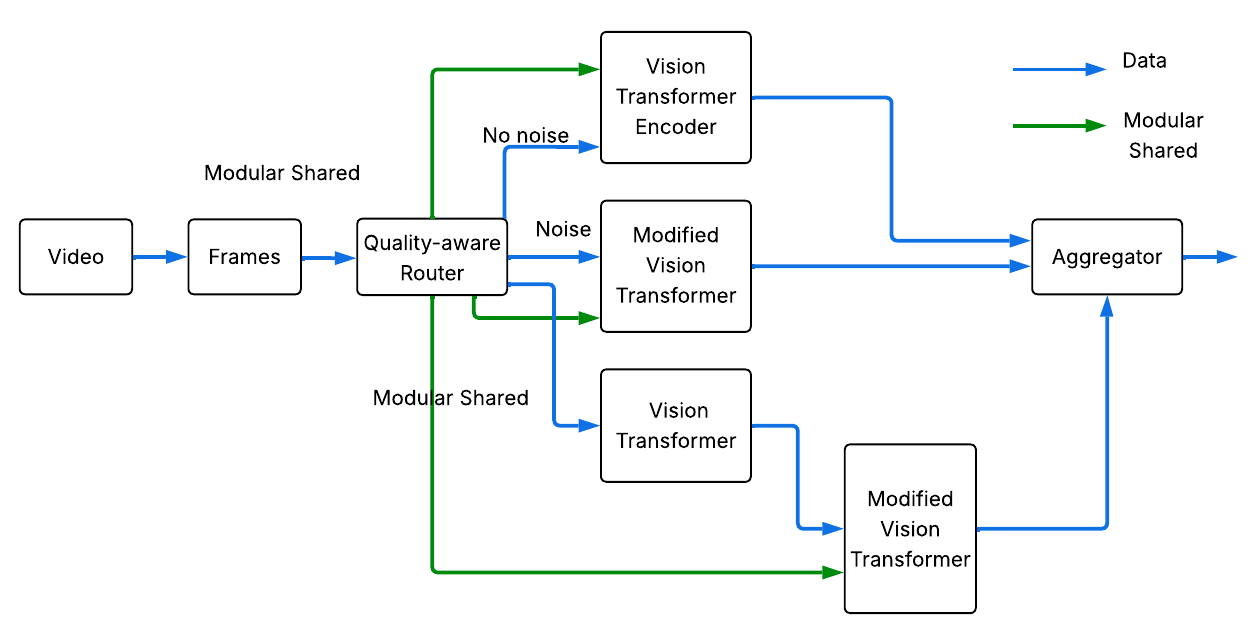}
    \caption{High-level architecture of the context-aware modular deep-learning approach.}
    \label{fig:approach}
\end{figure*}

\subsection{Overall Framework}
The modular deep learning framework \cite{pfeiffer_modular_2023} with model selection \cite{nascimento2018context} is illustrated in Figure \ref{fig:approach}. As drone images enter the framework, they are initially analyzed by a router that inspects the frames for noise and blur. If there is no noise or blur, the frame is directed to a vanilla vision transformer architecture for segmentation. In contrast, if there is noise, the frame is routed to a modified vision transformer encoder that employs Fisher Vectors instead of patch extraction. If there is blur, the frame is directed to a vanilla vision transformer encoder and subsequently to a modified vision transformer decoder that utilizes the unrolled Lucy-Richardson algorithm for deblurring. If both noise and blur are present, the frame is subjected to both the modified encoder and decoder. The final segmentation result is aggregated from one of the models and subsequently transmitted as output. By implementing modularity in the encoder, these functional components can be efficiently reused.

\subsection{Modular Routing}
The modular router facilitates model selection by inspecting frames for noise and blur. To ascertain whether a frame is noisy, the router compares the Mean Absolute Deviation of the image’s high-pass filtered coefficients with a predefined threshold and routes information to the denoise module accordingly. The routing function employs a threshold for the Laplacian of the image to route information to the deblur module. If both noise and blur are present in the frame, a model that incorporates both Fisher Vector encoding and a Lucy-Richardson-based decoder is selected.

\subsection{De-noising Module}
If there is noise, the data is sent to the modified vision transformer encoder that uses Fisher Vector instead of regular images patches. Incorporating Fisher-encoded patches will enhance the accuracy of video segmentation when integrated into a transformer architecture by effectively managing noise in the source data. 

Consider the video to be a set of image frames. This can be represented as: $X \in R^{T \times H \times W \times C}$
 , where T is the frame number, H is height, W is width and C is number of channels.

Vision transformer \cite{dosovitskiy_image_2021} first breaks the image into patches as follows:$x_p \in R^{N \times (P^2 \dot C)}$
 , where N is the number of patches of dimension PxP.

These patches, including the noise present in the data pass through multi-head attention step along with the position embedding:

\begin{equation}
z_0 = [x_{class};x_p^1E; x_p^2E...;x_pNE]+E_{pos}
\end{equation}

\begin{equation}
z'_l = MSA(LN(z_{l-1}))+z{_{l-1}}
\end{equation}

This is followed by Layer Normalization and MLP as shown below:
\begin{equation}
z_l = MLP(LN(z'_l))+z'_l
\end{equation}
\begin{equation}
y = LN(z_L^0)
\end{equation}

Fisher Vectors encode the data as a Gaussian Mixture of Models (GMMs). Data is assigned to each model through a “soft” probability that positions data strongly belonging to a model closer to its center and noisier data further away. By varying the learnable parameters, Fisher Vectors have been employed in traditional image processing algorithms for denoising purposes.

FV descriptor, 
\begin{equation}
\rho_n(k) = \frac {N(f_n|\mu_k,\sigma^2_k)v_k}{\sum_{j=1}^kN(f_n|\mu_j,\sigma^2_j)v_j}
\end{equation}

Instead of passing the entire noise patch for attention, the Fisher Vector can be calibrated to transmit only those features that are strongly associated with GMMs. This ensures that noises present in the patches will not disrupt the globally learned GMMs. Consequently, the transformer will be able to concentrate its attention on patches devoid of noise. This enhancement will augment the accuracy of the segmentation performed by the vision transformer.  

\subsection{De-blur Module}
If there is blur in the image, it will be sent to the normal vision encoder. Literature research suggests that the use of the Lucy-Richardson (LR) algorithm for deblurring images is effective.  A novel idea proposed is that by introducing an unrolled version of the LR algorithm as part of the transformer decoder, it can deblur the video and improve the accuracy of segmentation.

The LR algorithm iteratively estimates the deblurred image by comparing the predicted deblurred image in each iteration to the original deblurred image and deconvolving the difference between the two. This can be replaced by a unrolled version that can be integrated into the transformer decoder.

The following is the iterative step of a traditional LR algorithm:
{\small
\begin{equation}
f^{(k+1)}(x, y) = f^{(k)}(x, y) \cdot 
\left[ \frac{g(x, y)}{(f^{(k)}(x, y) * h(x, y))} * h_{\text{reversed}}(x, y) \right]
\end{equation}
}


For the unrolled version, the Point Spread Function (PSF) update would be:
{\small
\begin{equation}
h^{(k+1)}(x, y) = h^{(k)}(x, y) \cdot 
\left[ \frac{g(x, y)}{(h^{(k)}(x, y) * f^{(k+1)}(x, y))} * f^{(k+1)}_{\text{reversed}}(x, y) \right]
\end{equation}
}


In the proposed architecture, in the transformer decoder, with the image features as keys and values, an estimate of the PSF is used as a query. This makes the attention mechanism look at different parts of the image and adjust the PSF. This results in the deblurred image being fed to the final segmentation step that performs in a more accurate manner. This can be represented as follows:

\begin{equation}
  \text{Features}_{\text{input}}^{(k)} = \text{Concatenate}(f^{(k)}, h^{(k)}, g, f^{(k)} * h^{(k)})
\end{equation}

\begin{equation}
\begin{split}
f^{(k+1)} = \text{UpdateLayer}\big(
    \text{Features}_{\text{input}}^{(k)}, \\
    \text{TransformerBlock}_{\text{image}}(
        \text{PatchEmbed}(\text{Features}_{\text{input}}^{(k)}))
\big)
\end{split}
\end{equation}


\begin{equation}
\begin{split}
h^{(k+1)} = \text{UpdateLayer}_h\big(
    \text{Features}_{\text{input}}^{(k)}, \\
    \text{TransformerBlock}_{\text{psf}}(
        \text{PatchEmbed}(\text{Features}_{\text{input}}^{(k)}))
\big)
\end{split}
\end{equation}


Depending on the model to which the data was routed to, the segmentation results are aggregated to produce the final result.

\section{Experimental Setup}
This section describes the experimental setup, including the datasets, metrics, and evaluation procedures that were used.
The experimental setup is designed to investigate the hypothesis that the modular deep learning framework, which addresses blur and noise, will substantially enhance the performance of segmentation on the sorghum dataset \cite{genze_improved_2023}. 

\subsection{Dataset}
\begin{figure*}[htbp]
    \centering
    \includegraphics[width=0.8\linewidth]{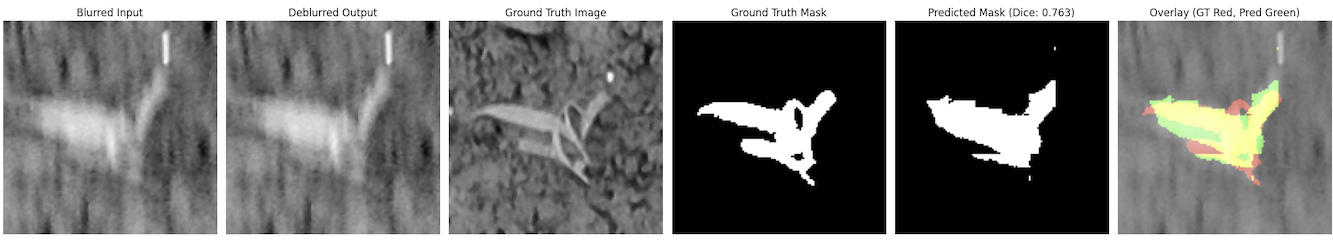}
    \caption{Example data with results.}
    \label{fig:data}
\end{figure*}

The sorghum dataset \cite{genze_improved_2023} comprises 1300 non-overlapping, blurred-sharp image pairs captured by a drone traversing a sorghum field. The ground truth comprises labeled segmentation data derived from the blurred and sharp images. Each blurred-sharp image pair, along with its associated segmentation result, is represented as four quadrants within a single file. Figure \ref{fig:data} illustrates the contents of a typical file. The first column is the blurred image, while the third column is the corresponding sharp image. The fourth column consists of the ground truth segmentation mask obtained for the sharp image that is used for training. The dataset was partitioned to allocate 80\% of the images for training and 20\% for validation.

\subsection{Metrics}
The experimental setup employs the same metric as the original authors \cite{genze_improved_2023} to facilitate systematic comparison of results. For the segmentation task, the authors employ the Sorenson-Dice coefficient, commonly referred to as the Dice-Score (DS). This can be represented as:
\begin{equation}
DS = \frac {2 \cdot TP} {2 \cdot TP + FP + FN}
\end{equation}

In the context of segmentation evaluation, True Positive (TP) represents the number of correctly predicted segmentation results that were indeed positive. False Positive (FP) denotes the number of negative results that were erroneously predicted as positive. Conversely, False Negative (FN) signifies the number of positive results that were incorrectly predicted as negative.  

\subsection{Hardware and Software}
The experiments were conducted on an Apple computer equipped with an Apple M4 Pro processor, 24GB of memory, and a 10-core GPU. The software utilized Python 3.12.10, along with pytorch, scikit-learn, and scikit-image, for its development.

\section{Results and Analysis}\label{sec2}

Figure \ref{fig:data} presents the output generated by this modular deep learning framework. The ground truth images and masks are displayed in columns one, three, and four, respectively. The de-blurred and de-noised image is showcased in column two. Column five depicts the predicted mask, while column six overlays the predicted (green) and ground truth (red) masks onto the blurred image. This visualization effectively demonstrates the visually strong alignment between the predicted and ground truth masks. Next, the results will be analyzed using quantitative metrics.

The dice scores from the experiment are presented in Table \ref{tab:results}. The original authors \cite{genze_improved_2023} reported a Dice-Score of 0.8373 for segmentation. When blurred images were segmented using a vanilla vision transformer without accounting for noise or blur, the Dice-Score was 0.7794. Employing the modular approach, the Dice-Score was enhanced to 0.8492, surpassing the results of the original authors. A more detailed ablation study is presented in the subsequent section.

\begin{table}[htbp]
\caption{Segmentation Results.}
\begin{center}
\begin{tabular}{|l|c|c|c|c|}
\hline
Metric & Genze & ViT & ViT+Modular Routing+FV+LR \\
\hline
Dice-Score & 0.8373 & 0.7794 & 0.8492 \\
\hline
Time (minutes) &  & 3.5 & 11 \\
\hline
Epochs &  & 75 & 75 \\
\hline
\end{tabular}
\label{tab:results}
\end{center}
\end{table}

\section{Discussion}

\begin{table*}[htbp]
\caption{Ablation Studies.}
\begin{center}
    \begin{tabular}{|l|c|c|c|c|c|c|c|}
    \hline
       Metric  & Genze & ViT & ViT+FV & ViT+LR & ViT+FV+LR & ViT+Modular Routing+FV+LR\\
       \hline
        Dice-Score & 0.8373 & 0.7794 & 0.7768 & 0.7492 & 0.8512 & 0.8492 \\
        \hline
        Time (minutes) &  & 3.5 & 10.5 & 43 & 50 & 11 \\
        \hline
    \end{tabular}    
\label{tab:ablation}
\end{center}
\end{table*}

Table \ref{tab:ablation} presents the ablation study conducted to evaluate the performance of various components within the modular deep learning framework for image segmentation. The first column displays the Dice-Score, which was demonstrated by the dataset creators \cite{genze_improved_2023}. The second column showcases the vanilla vision transformer performing the segmentation, resulting in a score of 0.7794 and requiring 3.5 minutes for training. 

Following the aforementioned modifications, the noise and blur reduction components were independently incorporated into the pipeline. The outcomes are presented in columns 3 and 4, respectively. Their performances were similar to the vanilla transformer. However, the training time increases without the presence of modular routing and quality based model selection.

Column 5 presents the results obtained by combining both the de-noise and de-blur modules. Notably, this combination achieves the highest Dice-Score of all the tested combinations, reaching 0.8512. This outcome arises from the fact that all the image quality improvements are performed regardless of their necessity. The increased Dice-Score is at the expense of a much longer computational time of 50 minutes for training. 

The final column underscores the versatility of the modular deep learning framework. The segmentation quality, as assessed by the Dice-Score, is 0.8492, surpassing the score achieved by the original authors but falling short of the score obtained when de-blur and de-noise were applied to all images. However, by selectively utilizing these models, only when required, the solution achieves a significant reduction in computational time, albeit with a slight compromise in the Dice-Score.

\section{Conclusion and Future Work}
This study presented a novel, modular deep-learning approach for weed segmentation in drone imagery. To address common challenges like noise and motion blur, the architecture was designed to route data through specialized pre-processing models before segmentation. A transformer-based encoder was used to effectively remove noise, and a Lucy-Richardson based decoder was used to correct motion blur. This modular approach not only improved the overall segmentation accuracy, as demonstrated by an increased Dice-Score, but also significantly reduced the computational time. This work offers a powerful and efficient solution for precision agriculture, enabling more accurate and timely weed management. Future work includes testing this solution on the field and extending the agricultural application to predict harvest.

\bibliographystyle{IEEEtran} 
\bibliography{sn-bibliography.bib}

\end{document}